\documentclass[letterpaper, 10 pt, conference]{ieeeconf}  %

\IEEEoverridecommandlockouts                              %

\usepackage{graphics} %
\usepackage{epsfig} %
\usepackage{mathptmx} 
\usepackage{times} 
\usepackage{amsmath} 
\usepackage{amssymb}  
\let\labelindent\relax
\usepackage{enumitem}

\title{\LARGE \bf
Person Transfer in the Field: Examining Real World Sequential Human-Robot Interaction Between Two Robots
}

\author{Xiang Zhi Tan$^{1}$, Elizabeth J. Carter$^{2}$, Aaron Steinfeld$^{2}$%
\thanks{*This work was funded by the National Institute on Disability, In-
dependent Living, and Rehabilitation Research (90DPGE0003).}%
\thanks{$^{1}$ Xiang Zhi Tan is with Northeastern University
        {\tt\small zhi.tan@northeastern.edu}}%
\thanks{$^{2}$ Elizabeth J. Carter and Aaron Steinfeld are with Carnegie Mellon University
        {\tt\small \{ejcarter@andrew., steinfeld@\}cmu.edu}}%
}

\begin{document}

\maketitle
\thispagestyle{empty}
\pagestyle{empty}

\begin{abstract}
With more robots being deployed in the world, users will likely interact with multiple robots sequentially when receiving services. In this paper, we describe an exploratory field study in which unsuspecting participants experienced a ``person transfer'' -- a scenario in which they first interacted with one stationary robot before another mobile robot joined to complete the interaction. In our 7-hour study spanning 4 days, we recorded 18 instances of person transfers with 40+ individuals. We also interviewed 11 participants after the interaction to further understand their experience. We used the recorded video and interview data to extract interesting insights about in-the-field sequential human-robot interaction, such as mobile robot handovers, trust in person transfer, and the importance of the robots' positions. Our findings expose pitfalls and present important factors to consider when designing sequential human-robot interaction.
\end{abstract}

\section{Introduction}
As more robots are being tasked with more complex human service scenarios, individual robots are likely not designed to complete all aspects of the task due to functionality trade-offs or service requirements. Similar to existing human-human interactions in sandwich shops or hospitals, users would likely interact with multiple different specialized robots sequentially to complete those tasks. In our prior work, we coined the term ``person transfer''~\cite{tan2021charting} to describe the act of transferring a user from one robot to another. Our work and the community have explored various aspects of this atomic interaction, including robot-robot communication~\cite{tan2019handover, williams2015covert, soderlund2021robottorobot} and spatial formation~\cite{tan2022group}.
\par
As a novel but fast-approaching class of interaction, there has been sparse work exploring how these interactions will occur and be perceived outside of a laboratory setting. Prior work has suggested that laboratory environments may lower perceptions of risk \cite{morales2019interaction} and heighten awareness of certain details of robot behaviors \cite{tan2018bully}.
\par
We conducted an exploratory field study to better understand how findings from the controlled user studies might translate to the real world. We believe that real conditions, such as having numerous people traversing the interaction environment and not having a scheduled appointment with the robots, can affect how people interact with them and provide insights not available in a laboratory environment. 
\par
In this study, the unsuspecting public interacted with a stationary robot that summons a mobile robot to deliver stickers to the users. We recorded various instances of these interactions and used our recorded data to extract important themes such as trust and group membership.

\begin{figure}
    \centering
    \includegraphics[width=\columnwidth]{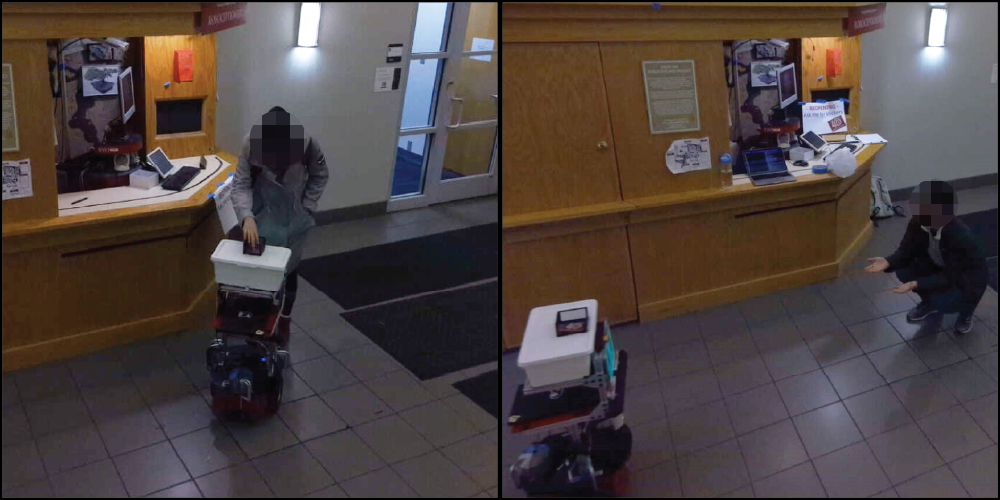}
    \caption{Left: A participant taking the stickers from the mobile robot as the mobile robot finished its movement. Right: A participant reacting to the arrival of the mobile robot.}
    \label{fig:teaser}
\end{figure}

\section{Related Work}
Human-Robot Interaction (HRI) researchers have long deployed robots in the world and observed how the public interacts with them. Prior work has explored robots being deployed on university campuses~\cite{gockley2005designing}, supermarkets~\cite{lewandowski2020socially}, hospitals~\cite{hebesberger2017long}, shopping malls~\cite{kanda2009affective}, and museums~\cite{nourbakhsh2003mobot, shiomi2006interactive}.
Rothenbucher~et~al.~\cite{rothenbucher2016ghost} created a wizard-of-Oz controlled autonomous vehicle and analyzed how pedestrians responded to the car's actions. Sun~et~al.~\cite{sun2023yousu} developed a public robot art display and investigated how it attracted the public to interact with it. Edirisinghe~et~al.~\cite{edirisinghe2024field} tested how an autonomous robot could encourage purchases at a hat store. Tuyen~et~al.~\cite{tuyen2023study} explored the importance of robot gestures when presenting information in a food ordering interaction in a cafe setting. Hauser~et~al.~\cite{hauser2023influencing} showed that pedestrians found a quadruped robot that display caine-like body language to be more friendly and likable.
\par
Besides deploying their robots, researchers have also investigated the public's reaction and opinions to commercial robots. Reig~et~al.~\cite{reig2018field} interviewed the public about their attitude towards autonomous vehicles deployed in their city. Han~et~al.~\cite{han2023robot} investigated the impact of sidewalk autonomous delivery robots on people with motor disabilities. Pelikan~et~al.~\cite{pelikan2024encountering} analyzed videos of autonomous delivery robots and explored the robots' interaction with a ``streetscape'' -- the people, objects, and interactions that happen on the street.
\par
However, there has been little work exploring how unsuspecting users interact with more than one robot. Shiomi~et~al.~\cite{shiomi2009field} described a field study in which two robots performed various tasks in a shopping mall. As part of a guidance task, the two robots coordinated their actions such that one robot would lead the guest to another robot that then welcomed the guest to a store. In a cross-cultural study, Fraune~et~al.~\cite{fraune2015three} investigated how the number of sociable trash box robots affects the interaction and people's perception of the robots in a cafeteria. They found participants responded more positively to a single social robot and a group of functional robots than to a group of social robots and a single functional robot. Prior work has also deployed ``Robot-Manzai'', a setup where two robots acted as passive social media and communicated with each other in front of bystanders with the goal of conveying information to bystanders in science museums~\cite{shiomi2006interactive} and train stations~\cite{hayashi2007humanoid}. Most prior work in the field focuses on other important aspects of human-robot group interaction and does not investigate how people react to sequentially interact with multiple robots.

\section{Method}
This study took advantage of an existing deployed robot on our university campus, the Roboceptionist. This is a social robot system started in 2003 as a long-term robotic platform, and it has been involved in multiple prior studies in HRI~\cite{gockley2005designing,makatchev2009incorporating,sabanovic2006robots}. Roboceptionist has undergone various changes throughout its deployment, most notably in its character and backstory. The latest version of Roboceptionist is an agent named ``Tank''. 
\par
To capture the nuances missed in laboratory settings and observe unique one-off situations from the combination of various factors in the environment, we used a qualitative observational study approach. We recorded instances in which people interacted with our robots and reviewed them for interesting factors of interactions and pitfalls with our system. After some interactions, we approached the participants and interviewed them about their experiences. Because participants were likely to only interact with Tank once due to novelty, all participants experienced the same scenario in which they received a sticker from the robots.
\par
This study was conducted in the field and passersby were recorded regardless of whether they interacted with the robot; thus, we took additional measures to ensure that the privacy of the participants was maintained. We placed disclosure signs on the edge of the recording area to inform participants they were being video and audio recorded. Passersby could also ask to remove their data using an online form. While potential participants were informed that they were being recorded, we did not mention the involvement of the robots. This study went through a full board review by Carnegie Mellon University's Institution Review Board and was approved.

\subsection{Study Environment \& System}
\begin{figure}
\includegraphics[width=\columnwidth]{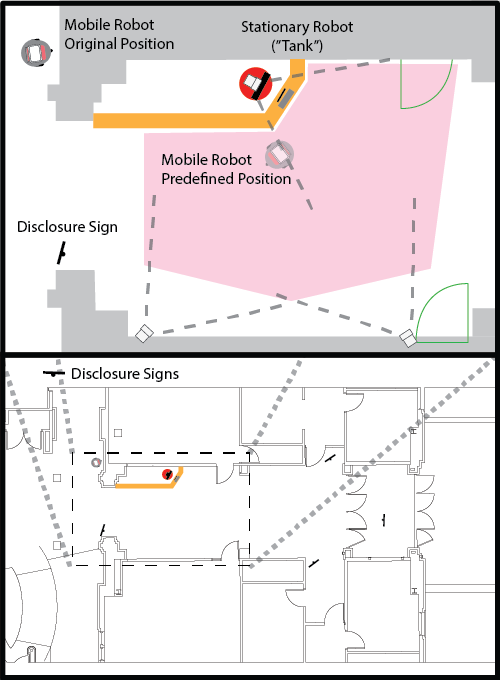}
\caption{Illustrations of the study scene layout. The bottom image shows the layout of the building with the locations of the disclosure signs. The top image shows the zoomed-in study location with the recorded region marked in pink.}
\label{fig:fs:layout}
\end{figure}
This study took place in the entrance hallway of Newell-Simon Hall on Carnegie Mellon University's Pittsburgh campus in the United States. A bird's-eye view illustration of the layout is shown in Figure~\ref{fig:fs:layout}. The Roboceptionist system sits in a wooden booth in the entrance hallway. A partial wall in front of the booth creates a physical barrier between humans and the robot. A screen and keyboard for user input are placed on a wooden ledge directly in front of the robot. A small empty box where the stickers could have been was placed near the keyboard. The Roboceptionist system is visible to visitors as soon as they enter the building from the main entrance. The second robot, a mobile robot, hid around the corner out of view and only appeared when summoned.
\par
The Roboceptionist system consists of a stationary robot body (iRobot B21R) and a screen mounted on a pan-tilt unit that acts as the head. %
We leveraged the Roboceptionist's ability to pan its head to convey gaze direction. The screen shows the animated face of ``Tank'', a muscular face that wears a headset that it uses to simulate taking phone calls. We also added a microphone and a speech-to-text capability to the system, which allowed users to communicate with Tank through spoken speech. The speech-to-text was disabled when the keyboard was in use. We augmented the environment with three Azure Kinect cameras that provided information about people's poses and locations.
\par
The mobile robot was a custom robot that had a Pioneer P3DX as its base. An aluminum structure was built on top of the base with a small box on top holding the stickers. The mobile robot uses a 2D lidar for navigation and has a tablet in the front as a face.

\subsection{System}
The system utilizes a combination of ROS 1~\cite{quigley2009ros} and $\backslash$Psi~\cite{bohus2017psi}. $\backslash$Psi handles the recording of video data and pipes the information to the ROS 1 system which controls the mobile robot and communicates with Tank's original code base. We build upon Tank's original code base~\cite{gockley2005designing, makatchev2009incorporating} and create an interface between its IPC framework and ROS 1. The mobile robot uses the ROS Navigation stack~\cite{marder2010office} with a lattice local planner~\cite{pivtoraiko2005generating}. When the study was active, Tank's system transitioned into a puppeteering mode and the behaviors of both robots were controlled by a Behavior Machine (i.e., a custom hierarchical state machine and behavior tree hybrid system)~\cite{tan2022thesis}. We authored the following scenario using the Behavior Machine.

\subsection{Scenario}
When the study was active, we enabled a special mode called ``sticker study''. In this mode, Tank told the participants interacting with it that it was giving out stickers as part of its reopening. If the participants indicated they wanted the stickers, Tank would inform them that it had run out of stickers and would summon another robot (``green mobile robot'') who had more stickers. The mobile robot would then drive around the corner and join the interaction. We piloted a group-joining algorithm that did not work most of the time due to technical errors and the busyness of the hallway; instead, the robot defaulted to a predefined position which was directly next to the robot and facing where the user would likely be.
\par
After arriving and exchanging greetings with Tank, the mobile robot prompted the participants to take a sticker from a top-mounted tray. The experimenter, who was standing nearby and out of the way, could remotely command the robots to skip the prompt if participants had already taken the stickers. Afterward, the mobile robot told Tank that someone would be coming to refill Tank's stickers. We added this brief conversation to observe how people would react and observe people's movements while a group interaction with the two robots was in progress. After the conversation, the mobile robot informed the participant that it had to leave, and it departed. Tank then looked at the participants and told them about the study, mentioning that they could approach the nearby experimenter if they had any questions.

\section{Study Setup}
\subsection{Participants \& Recruitment}
Our study included four types of participants:
\begin{description}
    \item [Passersby] -- 
    These were people who passed by our robots without interacting with the robots or observing any human-robot interactions.
    \item [Observers] --
    These were people who passed by and observed the robots interacting,  e.g., by slowing down or stopping to watch someone else interact with them. However, they did not directly interact with the robots.
    \item [Participants Group A] --
    These were people who took part in some or all of the multi-robot interactions. Some participants left the interaction halfway.
    \item [Participants Group AA] --
    These were people who were in group A and also answered a few questions that the experimenter asked. This interaction took less than 5 minutes. Participants in this condition were not compensated.
    \item [Participants Group AB] --
    These were people who were in group A but also participated in a 15-minute interview and completed a questionnaire after the interaction. Participants were compensated \ USD 10 for their time.
\end{description}
After participants in group A completed the interaction, the experimenter approached the participants and asked if they had any questions before informing them about the interview. We were not able to intercept all participants; some left the building while we were supervising the mobile robot as it drove back to its starting point.  Furthermore, not all participants agreed to be interviewed due to time commitments. We also did not interview participants who knew or recognized the experimenter. There were two groups of approached participants (AA and AB) because we wanted to provide the options to answer a few questions (AA) or complete a 15-minute interview (AB).
\par
We recorded 7 hours and 2 minutes of complete study data (e.g., video data, robot state, etc.) over 4 weekdays. All sessions took place in November 2021 between 11:20 am and 6:30 pm. We only enabled the microphone input system for some sessions. We observed during pilot testing that the vast majority of people who walked through the hallway ignored Tank. Therefore, we added a sign in front of the Roboceptionist system to advertise that it was giving away free stickers.
\par
In our post-study analysis, we observed 18 person-transfer interactions involving at least 40 people. We obtained this number by counting the number of people in front of Tank when the mobile robot was summoned. We believe this is a lower bound: there were multiple instances in which people approached or interacted with the robots after the mobile robot arrived. We excluded instances where people did not successfully request the stickers or just stared at Tank. We interviewed 11 people (3 in group AA and 8 in group AB).

\section{Findings \& Discussion}
For this exploratory field study, we took an exploratory approach in our analysis. We watched the recordings of the 18 interactions and used the interviews to extract important themes.

\subsection{Overall Impressions}
Overall, participants told us that they thought the interactions were ``cool'', ``fun'', and ``really neat''. For most participants, the highlight of the interaction was the arrival of the mobile robot. This was also reflected in our observations in which participants expressed excitement when they saw the mobile robot approaching them.

\subsection{Trust in Person Transfers}
\begin{figure*}[ht]
    \centering
    \includegraphics[width=\textwidth]{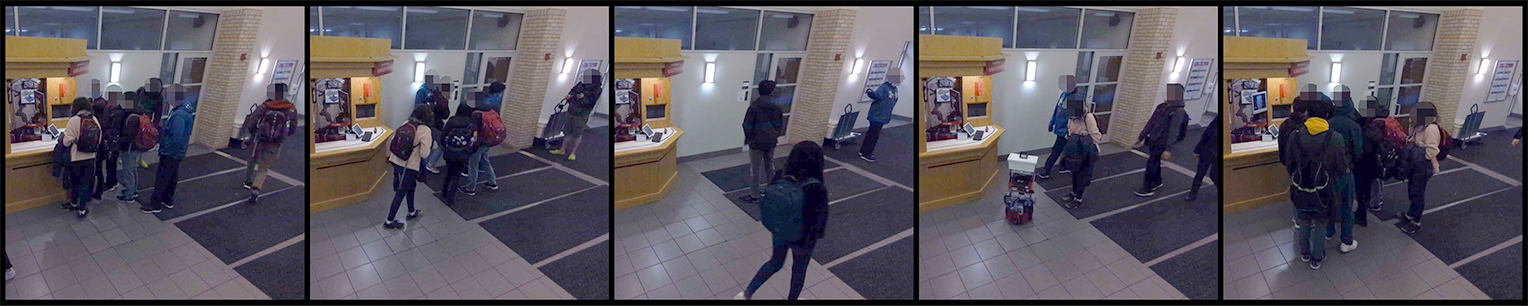}
    \caption{Video stills showing participants leaving after Tank informed them it had run out of stickers and returning after the mobile robot arrived. (1) A group of participants interacting with Tank. (2) The group walking away after Tank told them it ran out of stickers. (3) One of the participants (in blue) saw the mobile robot, pointed at it, and called for the group. (4 \& 5) The group returned and surrounded the mobile robot to get the stickers.}
    \label{fig:fs:leave-join}
\end{figure*}
One of the common themes brought up by the interviewed participants was that they were surprised that the mobile robot showed up with more stickers. Participants told us that they thought the sticker was a lie and that Tank was joking in saying that another robot would come to give out stickers. Their rationales were that the joke fit Tank's personality and that they had never seen another robot in the area before. Participants stated they did not believe it was true until they saw the mobile robot turning the corner. This reinforces the current novelty of interacting with multiple robots and the need for more research on this topic.
\par
Participants' behaviors confirmed this disbelief in the promise of another robot: in several interactions, after Tank informed a participant that it had run out of stickers, the participant(s) left the interaction. As the mobile robot moved towards Tank, the participants turned around and reengaged with the robots (Figure~\ref{fig:fs:leave-join}). Some participants intercepted the mobile robot and took the stickers from it as they left.

\subsection{Mobile Robot Handover}
The action of picking up the sticker from our mobile robot can be viewed as a ``handover'' of an item from our mobile robot to participants. In $13$ out of $18$ person transfer instances, participants took the stickers while the mobile robot was still trying to reach its final position. In the remaining $5$ scenarios, participants only took the stickers after being prompted by the mobile robot. In $4$ out of $5$ scenarios, the mobile robot positioned itself further away from the participants (an example of the distance is shown in Figure~\ref{fig:fs:other-people}). In the remaining one scenario, the participants were recording the interaction and did not immediately pick up the stickers.
\par
These results provided insights into how people perceive receiving items from a robot during handovers. The act of receiving an item requires the person to decide when to approach and take the item. In most cases, as the robot got close to the participants, participants took that as permission to take the items. In cases where it was further away, participants waited for the signal from the robot that it had completed its movement before taking the item.

\subsection{Change in Group Membership}
\begin{figure*}[ht]
    \centering
    \includegraphics[width=\textwidth]{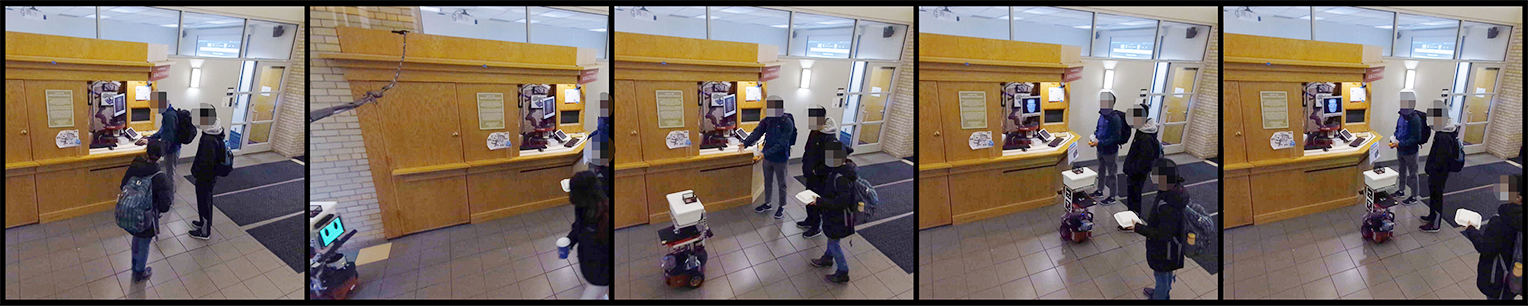}
    \caption{Video stills showing a participant (P4, blue backpack) joining an ongoing interaction. From left to right: (1) P4 joined two other participants who were interacting with the robot. (2) The mobile robot was summoned and moved towards them. (3) P4 stepped back as the robot approached them.  (4) P4 observed the interaction between the mobile robot and the other two participants. (5) P4 stepped further away as the interaction progressed.}
    \label{fig:fs:change-in-interactors}
\end{figure*}
Among the 18 sessions, we observed a few instances where someone joined an ongoing interaction. In one case, one participant (P4) joined two others who were already interacting with Tank. The first two participants requested the stickers and the mobile robot approached the group, moving to its predefined position. As the mobile robot got closer, P4 stepped back and moved away from the interaction. From the recording, we observed that P4 stood farther away and even stepped back as the two robots interacted with each other.
\par
In the post-interaction interview, P4 mentioned that it was unclear to them if they were part of the interaction because it was their friends who were initially interacting with the robot. As the mobile robot moved towards them, P4 was unsure if the mobile robot knew they were part of the group and moved out of the way. This sequence of interaction shows that the joining behavior has the potential to influence people's perception of group membership. A better, socially appropriate position could have made P4 feel confident they were part of the group and have been less likely to prompt them to move away.
\par
In one of the sessions, we observed a participant who stood to the side observing the interaction between the robot and another group of participants. Once the participant observed the group taking the stickers, they stepped in, took a sticker, and left. It was unclear if the person knew anyone who was initially interacting with the robots. 

\subsection{Mobile Robot Joining Position}
While the mobile robot moved to the predefined position next to the stationary robot in the majority of the interactions, we still collected valuable feedback on the position choice. Participants generally found the chosen position to be appropriate. One of the participants who experienced the predefined position stated that they wished the robot was closer. They talked about how while the position of the robot was where they expected a person to be, they believed the robot needed to come closer because it lacked the manipulation capabilities to hand over the object like a human would. Because the participant had to lean forward and take the object, they talked about how the robot should be only ``one hand'' (arm's length) distance away compared to the ``two hands'' distance that they experienced. 
\par
These findings, together with our observation of handovers and changes in groups, demonstrated the importance of a task-aware and human-aware joining strategy. A static position may lead to the system accidentally excluding others in the group, lead to people misinterpreting the robot's action, and likely be a poor position for certain tasks.  

\subsection{Effects of Keyboard Inputs \& Failures of Speech-to-Text}
When we first designed this study, one concern we had was that the keyboard input would limit the movement of participants to their starting position (where the keyboard was) to provide input. We ran this study with both keyboard input only and a combination of keyboard input and a microphone.
\par
Due to the combination of ambient sounds, hallway acoustics, and COVID-19 masking guidelines at the time of our study, our speech-to-text system was unreliable. We observed multiple participants who first attempted to use the speech system before stepping forward and interacting with the robot through the keyboard. In the interviews, participants reinforced our observations and discussed how the microphone was unreliable and they ended up using the keyboard.
\par
We also found some evidence that participants would move back to interact with the first robot. We observed participants moving back to the keyboard to type responses such as ``Thanks for the stickers'', ``got the sticker'', `'thanks''. This supports our intuition that the keyboard anchors the human's position during the interaction and changes the spatial dynamics of the interaction. Future work should explore how these anchors influence user positions and movements during person transfers and multi-robot interaction. For example, a factory worker using a fixed tool may choose to stay with the first robot instead of moving.

\subsection{Robot-Robot Communication}
After the participant picked up the sticker, the robots had a quick conversation. Participants had mixed reactions to the exchange. In a few cases, participants left after taking the stickers and did not wait for the interaction to finish. This was understandable as the conversation did not add any value to the service. The majority of the participants waited for the robots to finish their conversation. When asked about the conversation, one of the participants mentioned that it was a good addition as it showed that the robots could communicate and were on the same team. Some participants also stated that it was an artificial and performative act. One participant mentioned that they were surprised that the mobile robot could talk at all.

\subsection{Other People in the Scene}
\begin{figure*}
    \centering
    \includegraphics[width=\textwidth]{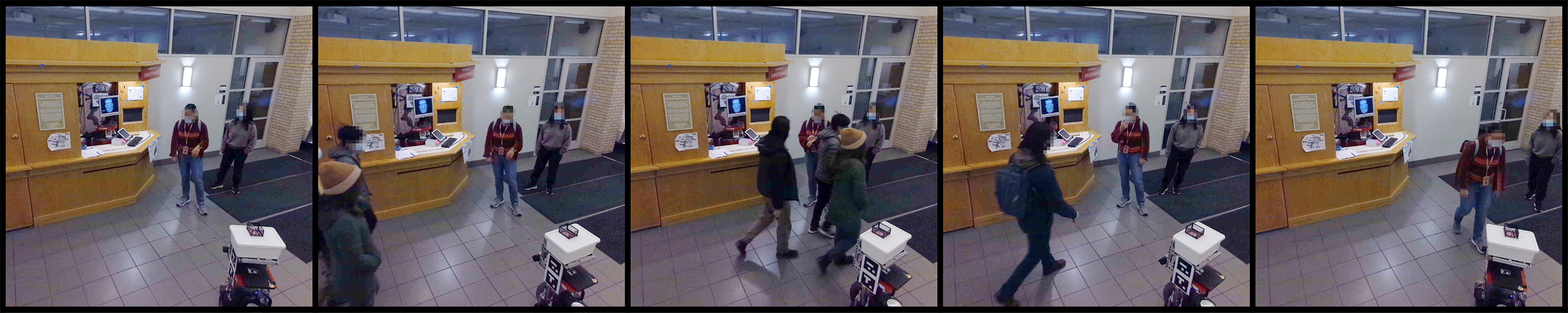}
    \caption{Video stills of a group of people walking through the group interactions. From left to right: The participant (red shirt) observed the greeting between robots. A group of people simply walked between the robot and the participants. After the group passed, the participant approached the mobile robot to get the sticker.}
    \label{fig:fs:other-people}
\end{figure*}
We were also interested in how bystanders and others in our study area interacted with our robots. In the majority of the sessions, when the mobile robot was close to Tank, we observed people moving around the robots and the participants interacting with our robot.
When there was a big gap between the participant and the mobile robot, we observed that most people in the passersby category simply walked through the gap, violating the human-robot group space (Figure~\ref{fig:fs:other-people}). However, we did observe one instance where a person consciously walked around the mobile robot even when the mobile robot was far away from the participant.
\par
\begin{figure}
    \centering
    \includegraphics[width=0.7\columnwidth]{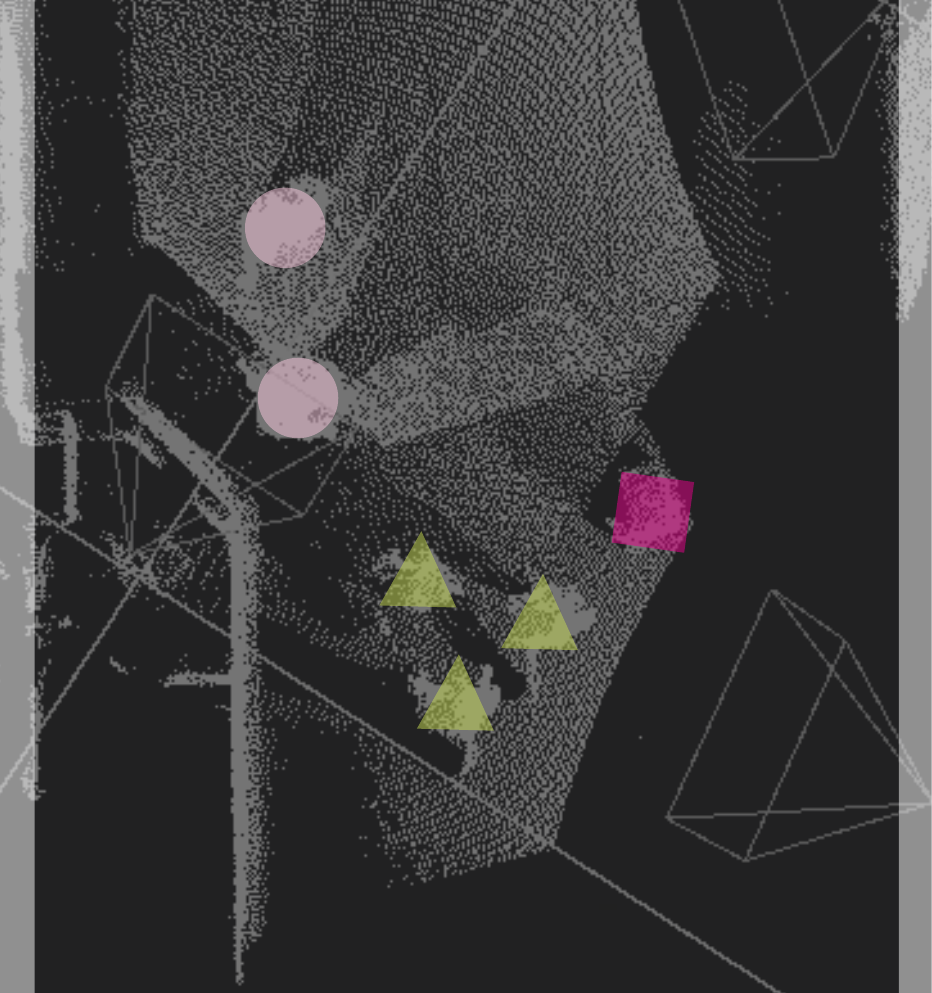}
    \caption{The top-down annotated point-cloud view of the scene. We annotated the positions of the participants (circle), the walls (white edges), the mobile robot (square), and the bystanders moving through (triangle).}
    \label{fig:fs:point-cloud-gap}
\end{figure}
It was unclear how much the layout of the hallway affected whether bystanders decided to walk through the gap. In the scene shown in Figure~\ref{fig:fs:other-people}, the mobile robot moved to the middle of the hallway and slightly blocked the default route through the hallway (as shown in Figure~\ref{fig:fs:point-cloud-gap}). While there was sufficient space behind the mobile robot for people to move through, it required large trajectory changes, and the space behind the robot was unlikely to fit a group of people. The function of the hallway as a means of moving between spaces may have also led bystanders to believe it was socially acceptable to violate the robots' and participants' O-space (the space between the participants in a group social interaction)~\cite{kendon1990conducting}. However, without interviewing the bystanders, it is unclear if they perceived the robot as being in a group with the two participants. 
\par
We also encountered situations in which other people in the scene purposely blocked the movement of the robot. As the hallway was filled with university students, we believed they were trying to test the capabilities and limitations of the robot. These curiosity-driven impedances are often observed when the robots are first being deployed~\cite{nomura2015children}.

\section{Limitations \& Future Work}
As pointed out above, group memberships and participants in the interactions were not static and constantly changed as people joined and left the interaction. We observed cases where up to 7 people concurrently interacted with our robots and cases where people left and rejoined the group throughout the interaction. A simplistic model to determine join positions such as those presented in~\cite{tan2022group} is likely insufficient. Future work should explore how to model group membership and incorporate that into mobile robot position selection.
\par
Our participants are not representative of the wider population. They were likely computer science students who were trying to get from one classroom to another. Furthermore, participants have seen Tank before, and we had to put extra signage to attract participants to interact with our robots. While our work is the first step in understanding person transfer in the field, future work should explore different tasks (e.g., guidance, food serving) and locations (e.g., hospitals, transportation hubs).

\section{Study Implications}
Combining our observational findings and interviews, we propose the following insights to consider when developing ``Person Transfer'' in the field:
\begin{enumerate}
    \item 
    In remote person-transfer scenarios (where the second robot is not co-located with the first robot)~\cite{tan2021charting}, the first robot should communicate clearly the existence of the second robot and its expected arrival time to instill confidence and trust.
    \item Ensure the second robot's joining trajectory and position respect the existing group formation to avoid alienating group members.
    \item For object handover, the second robot's position should be about one arm's length away from the participant. The robot can also use its positioning to signal whether or not an object is ready to be taken.
    \item While robot-robot communication that mimics social norms may be preferred~\cite{tan2019handover}, designers should account for situations in which participants skip pleasantries after receiving their services.
\end{enumerate}

\section{Conclusion}
Our field study expanded our understanding of how person transfer works and is perceived in the field. Our findings demonstrated why a context-aware socially appropriate mobile robot joining strategy is needed. A fixed position strategy will likely be less preferable for some tasks, alienate certain group members, and not react to the changes in group membership. Similarly, a bad, improper position can also lead to interruption by others as they walk through the interaction space. We also found that it is important for the first robot to communicate clearly how and when a transfer is going to happen. Furthermore, the dialog by Tank (``I ran out of stickers. Let me call green robot with more stickers'') likely did not instill confidence as it did not convey when and how the second robot would arrive. This work exposed pitfalls and generated insights on how to design sequential human-robot interaction in the field. Our findings highlighted the complexity of this space and future work should continue to explore how autonomous robotic systems can address the challenges we raised.

\section*{Acknowledgment}
We thank Reid Simmons and Greg Armstrong for helping us set up and use the Roboceptionist system. We also thank Prithu Pareek, Jodi Forlizzi, Selma Šabanović, the participants, and members of the Transportation, Bots, and Disability Lab at CMU.

\bibliographystyle{IEEEtran}
\bibliography{reference}

\end{document}